\documentclass[11pt]{article}

\usepackage[english]{babel}
\usepackage[numbers]{natbib}
\usepackage{physics}
\usepackage[letterpaper,margin=1in]{geometry}
\usepackage{amsfonts} 
\usepackage{multirow}

\usepackage{amsmath}
\usepackage{graphicx}
\usepackage{algorithm}
\usepackage{algpseudocode}
\usepackage{authblk}
\usepackage[labelfont=bf]{caption}
\usepackage{subcaption}
\usepackage{multirow}
\usepackage[colorlinks=true,urlcolor=cyan]{hyperref}
\usepackage[capitalize]{cleveref}

\makeatletter

\renewcommand\AB@affilsepx{, \protect\Affilfont}
\renewcommand\Affilfont{\small}
\makeatother

\title{Variable Length Embeddings}
\author[1,3]{Johnathan Chiu}
\author[2,3]{Andi Gu}
\author[3]{Matt Zhou}
\affil[1]{\href{https://runwayml.com}{Runway}} \affil[2]{Harvard University} \affil[3]{University of California, Berkeley}
\date{}

\usepackage{titling}
\begin{document}
\maketitle

\vspace{-1cm}
\begin{abstract}
In this work, we introduce a novel deep learning architecture, Variable Length Embeddings (VLEs), an autoregressive model that can produce a latent representation composed of an arbitrary number of tokens. As a proof of concept, we demonstrate the capabilities of VLEs on tasks that involve reconstruction and image decomposition. We evaluate our experiments on a mix of the iNaturalist \citep{inaturalist} and ImageNet \citep{imagenet} datasets and find that VLEs achieve comparable reconstruction results to a state of the art VAE, using less than a tenth of the parameters.
\end{abstract}

\section{Introduction}

We introduce a novel deep learning architecture, called Variable Length Embeddings (VLE). A VLE is an autoencoder that differs from traditional ones in one key aspect: whereas conventional autoencoders have a fixed embedding dimension, VLEs (as their name suggests) use a variable-length embedding dimension. 

Allowing the embedding dimension to vary is a natural idea: not all images are created equal. Images that contain more complex semantics should naturally require more resources to represent efficiently. Viewed through the lens of information theory, this is a well-known idea: we ought to use less resources to represent `easy'
samples. This idea is formalized by Shannon coding, which is an efficient compression scheme that maps samples $x$ to code words with length $l=\lceil -\log p(x) \rceil$. Here, the difficulty of a sample is measured by $-\log p(x)$, which means that we take frequently occurring samples to be easy. VLEs borrow from this idea, but diverge from the information-theoretic approach in terms of what is used to measure complexity. Rather than taking a bottom-up approach (which would involve modeling the density $p(x)$), we take a top-down approach. VLEs attempt to decompose the image into a sequence of semantically distinct objects, ordered by contextual significance. For images with sparse or simple content, the image should be accurately modeled with just a few tokens, whereas images with complex scenes may take much more. This iterative approach to reconstruction is again a well-known idea in different fields. In physics and applied mathematics, one often tries to represent an unknown function $f$ as a power series expansion (known as a perturbative expansion):
\begin{equation}
    f = f_0 + \epsilon f_1 + \epsilon^2 f_2 + \ldots, \label{eq:pert}
\end{equation}
where $\epsilon \ll 1$ is some small parameter. Put simply, $f_0$ represents the coarsest approximation to $f$. The addition of $\epsilon f_1$ corrects the coarse approximation $f_0$ by taking into account some additional details in $f$, then $\epsilon^2 f_2$ further improves this approximation, and so on. Similarly, in signal processing, one is often interested in finding a representation of a signal in terms of a weighted sum 
\begin{equation}
    f = \alpha_0 f_0 + \alpha_1 f_1 + \alpha_2 f_2 + \ldots. \label{eq:matching}
\end{equation}
The matching pursuit algorithm \citep{pursuit1993} is an effective, albeit suboptimal, method that finds this representation by iteratively `matching' $f_0,f_1,\ldots$ to the signal, at each iteration taking whichever function $f_i$ has the largest inner product with the remaining unmodeled components of the input signal.

\paragraph{Related works} As hinted at above, a variable length encoding is a natural idea for a number of tasks, one of which is compression. Indeed, \citet{toderici_full_2017} developed this idea using a long-term memory (LSTM) model to generate variable length codes to represent a given image. However, their focus is on achieving a maximal possible compression rate (i.e., faithfully representing an image using as few tokens as possible). In this work, we focus on using the variable-length approach as a means to find useful (i.e., interesting or interpretable) decompositions of the image. 
Other works such as DRAW \citep{pmlr-v37-gregor15} or diffusion models \citep{ho2020denoising} approach generative modeling by iteratively adding detail at different scales. We find that by focusing purely on generative modeling, although these models often produce interpretable results, they are unable to generalize to other downstream tasks, such as classification or captioning. On the contrary, our aim when designing VLE was that tokens could be used for any number of downstream tasks, including generative modeling, classification, or image captioning for future works.

\section{Methodology}

\subsection{Autoregressive Encoding}
Inspired by the decompositions of \cref{eq:pert,eq:matching}, we aim to represent images as a series of tokens, with each token being some vector in $\mathbb{R}^d$. The core of our idea is to use an autoregressive approach to generating these tokens; we propose a simple formulation for this. We have a single encoder $\mathcal{E}$ and decoder $\mathcal{D}$, and at each iteration the encoder input is equal to the remaining portion of the image that is unaccounted for by previous tokens. That is, the input at every iteration is the current residual (see \cref{alg:cap}). For training purposes, we set a maximum number of tokens $n_{max}$.

\begin{algorithm}
\caption{VLE Autoregressive Loop}\label{alg:cap}
\begin{algorithmic}
\State $n \gets 0$
\State $\hat{X}_n \gets 0$    \Comment{Initialize the reconstruction}
\While{$n < n_{max}$} 
    \State $z_n \gets \mathcal{E}(X - \hat{X}_{n-1})$
    \State $\hat{X}_n \gets \hat{X}_{n-1} + \mathcal{D}(z_n)$
    \State $n \gets n + 1$
\EndWhile
\end{algorithmic}
\end{algorithm}
One might naively define a loss to be the mean squared error between the final reconstruction $\hat{X}_{n_{max}}$ and the input $X$. However, limiting the number of tokens to $n_{max}$ is somewhat artificial (it is necessary as a practical matter for training purposes)\footnote{Future tasks could include determining the number of tokens needed per image through a metric or a learnable method.}, so it is rather unnatural to define a loss that depends on $\hat{X}_{n_{max}}$ alone. Our true objective is not merely to find a good final reconstruction; it is that each subsequent token improves the reconstruction as much as possible. Therefore, we might consider a loss in the form
\begin{equation}
    \mathcal{L} = \frac{1}{n_{max}}\sum_{n=1}^{n_{max}}MSE(X, \hat{X}_n).
    \label{eqn:mse-loss}
\end{equation}
This penalizes poor intermediate reconstructions, which encourages each token to play a significant role in modeling the image. We will term models trained with \cref{alg:cap} and a loss \cref{eqn:mse-loss} `vanilla VLEs', in contrast to a modified variant that we discuss in \cref{sec:sem-pix}.

\begin{figure}[H]
    \centering
    \includegraphics[width=\textwidth]{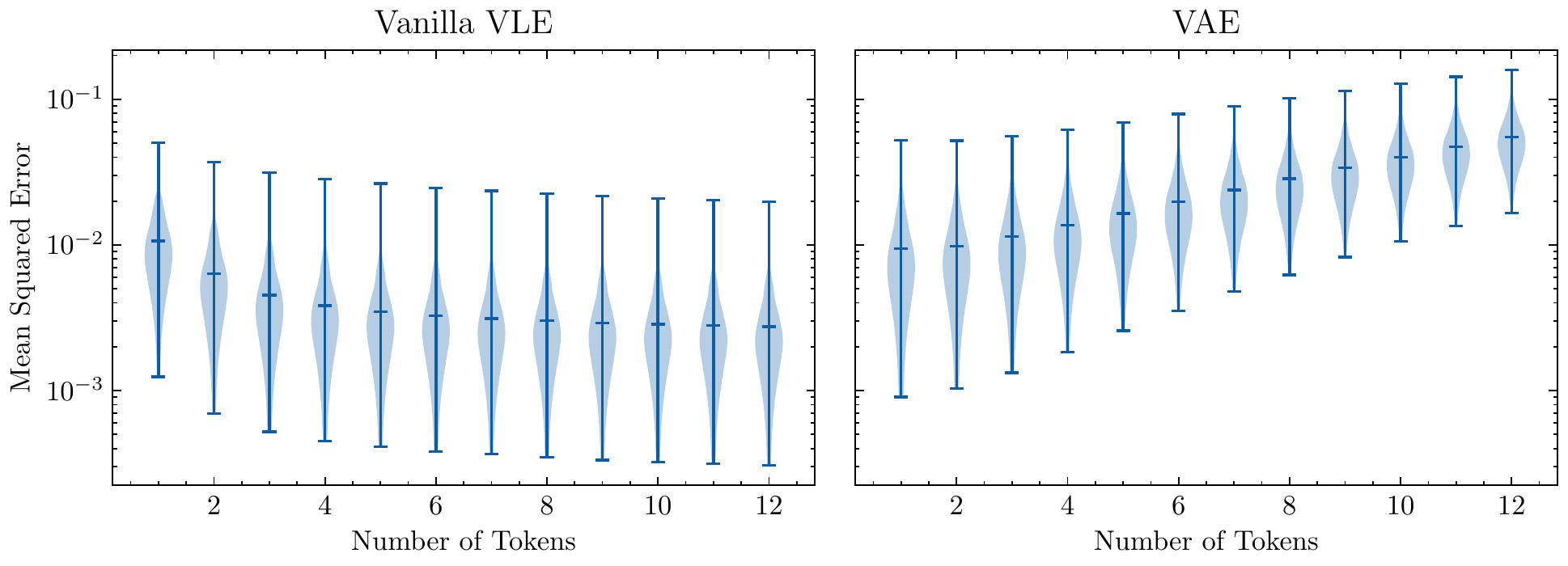}
    \caption{Applying \cref{alg:cap} with a trained VLE and a pretrained VAE, with an identical token dimension. The violin plot shows the distribution of mean squared error as a function of number of tokens used to model the image. This evaluation is done on the INaturalist and ImageNet validation set. Note the good generalization of the vanilla VLE: we only train the VLE with 4 tokens, yet the mean modeling error is still strictly decreasing past 4 tokens.}
    \label{fig:msextokens}
\end{figure}

Given the structure of \cref{alg:cap}, it may seem reasonable to use pretrained models for the encoder $\mathcal{E}$ and decoder $\mathcal{D}$. However, we find that the autoencoders do not generalize well without further training on $n_{max} > 1$ tokens (see \cref{fig:msextokens}). This is likely due to the fact that for later tokens $n \geq 2$, the encoder-decoder pair is being run on out-of-distribution inputs.

\subsection{Semantics vs. Pixels}\label{sec:sem-pix}
\begin{figure}[H]
    \centering
    \includegraphics[width=0.7\textwidth]{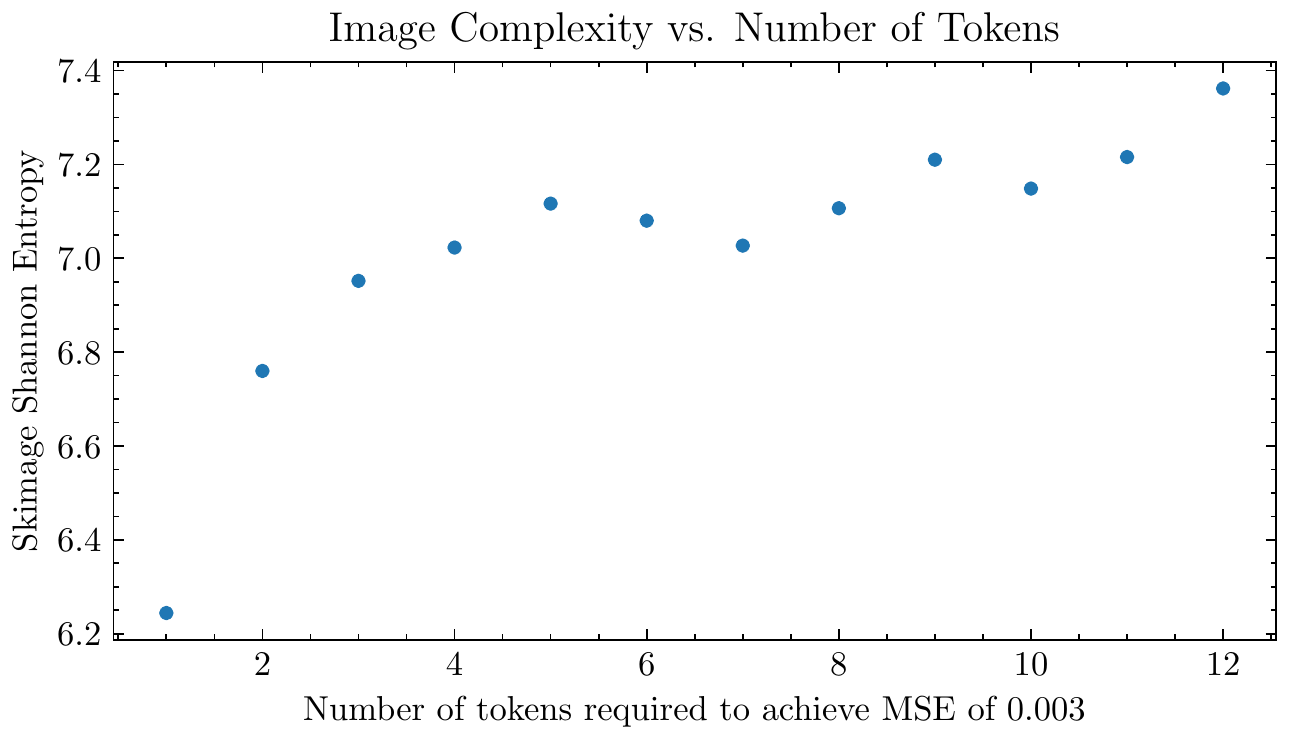}
    \caption{We show the number tokens required to reach a certain MSE threshold against the internal Shannon entropy of an image (computed using Scikit-Image's Shannon Entropy function \citep{scikit-image}). The plot shows an overall upward trend meaning that the as the image entropy increases, the number of tokens required to reconstruct the image increases as well.}
    \label{fig:entropy_plot}
\end{figure}

Losses in the form of \cref{eqn:mse-loss}, by design, encourage the model to simply match the \textit{pixels} of the input image as closely as possible. Indeed, models trained with this loss perform very well on reconstruction (for details see \cref{sec:eval,sec:vanilla-vae-res}). In fact, one can test a VLE model's dependence on pixel values by finding the number of tokens required to model an image to a given error, as a function of the entropy in the image's pixel distribution. The positive correlation between the two for vanilla VLEs (see \cref{fig:entropy_plot}) indicates that these models have a strong dependence on the complexity of the pixel distribution, rather than the semantic content of the image. 

Although this dependence on pixel distribution may be desirable in certain cases, we prefer a reconstruction that matches the \textit{content} of the image as closely as possible. In fact, this goal has led to the development of perceptual losses such as LPIPS \citep{zhang2018}. Unsurprisingly, we find that if we simply minimize \cref{eqn:mse-loss}, although we get faithful reconstructions, intermediate tokens often do not represent semantically distinct objects, and the model instead learns fairly elementary decompositions of the input (see \cref{sec:vanilla-vae-res}). For instance, a fairly typical mode was for early tokens to represent low frequency data in the image, and later tokens to contain higher frequency data. Although this may be interesting in its own right, we prefer a model that is able to model semantically different objects with each token, since representations of an image that use different tokens for different objects are more:
\begin{itemize}
    \item useful for downstream tasks, such as classification or caption generation,
    \item intepretable,
    \item and amenable to generative modeling.
\end{itemize}

We attempt to remedy this by imposing a loss which encourages each token be distinct from all other tokens. This stems from the observation that semantically distinct objects are typically spatially localized, whereas simpler decompositions (such as a frequency-space decomposition) often reconstruct the image globally. We represent this distinctness loss as 
\begin{equation}
    \mathcal{L} = \exp(-MSE(\mathcal{D}(z_n), \hat{X}_{n - 1})).
    \label{eqn:diff-loss}
\end{equation}
This loss is minimized when the intermediate reconstruction on token $n$ is maximally different from the sum of the previous $n - 1$ reconstructions.\footnote{In our initial approach we applied Binary Cross Entropy (BCE) and found this loss to be numerically unstable, as it can diverge. The loss in \cref{eqn:diff-loss} has an inherent benefit such that the loss is maximized at $1.0$ and can only occur if each reconstruction is equivalent to the sum of the previous reconstructions.} However, imposing this loss alone pushes the model to decompose the image into a color scheme representation (where different color channels are modeled for each token). In light of this, we move one step further and introduce a `mask' component which we output from the initial layers of the encoder. In doing so, we provide a guidance mechanism to the encoder detailing an area/region of the image to encode. Rather than imposing the loss in \cref{eqn:diff-loss} on the reconstructions, we impose it on the \textit{mask} instead:

\begin{equation}
    \mathcal{L}_{mask,n} = \exp(-MSE(\tilde{M_n}, \hat{M}_{n - 1})).
    \label{eqn:diff-loss-mask}
\end{equation}

\begin{figure}
    \centering
    \includegraphics[width=\textwidth]{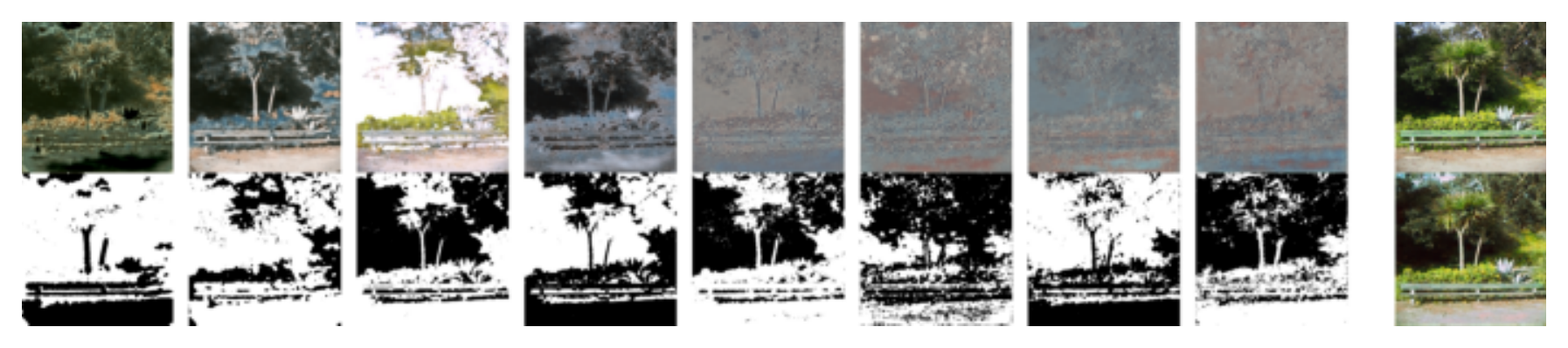}
    \includegraphics[width=\textwidth]{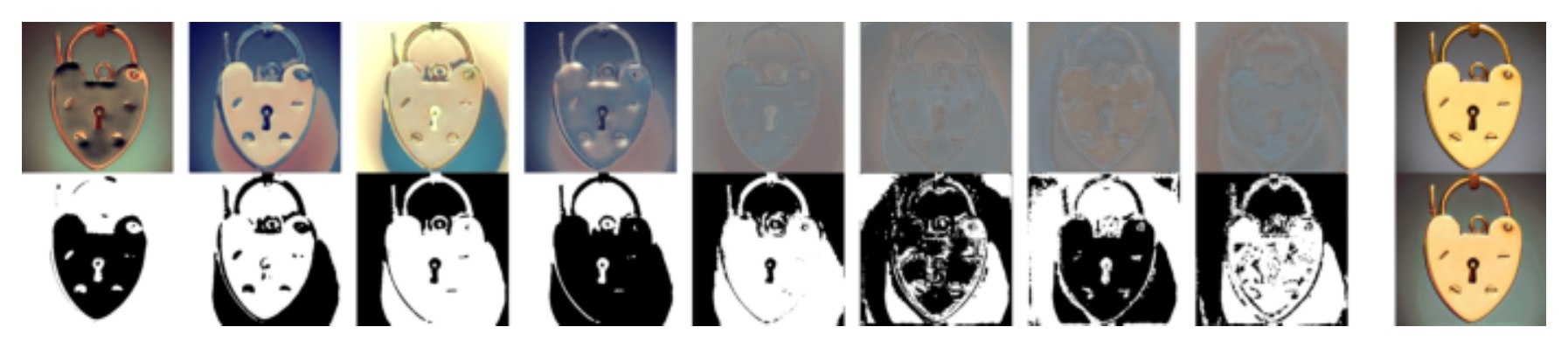}
    \caption{Intermediate reconstructions (top row) and corresponding masks (bottom row). The final column is the source (top image) and reconstruction (bottom image). We emphasize here that these masks are produced using self-supervision. See more results in \cref{sec:mask-vae-res}.}
    \label{fig:in-doc-examples}
\end{figure}

We additionally modify the loss represented in \cref{eqn:mse-loss} to incorporate the information of the mask. We define this as such:
\begin{equation}
    \mathcal{L}_{rec,n} = \frac{1}{D}||\tilde{M}_n \odot (X - \hat{X}_n)||_2^2.
    \label{eqn:mse-loss-with-mask} 
\end{equation}
Combining \cref{eqn:diff-loss-mask} and \cref{eqn:mse-loss-with-mask} results in our final loss defined as such:
\begin{equation}
    \mathcal{L} = \frac{1}{n_{max}} \sum_{n=1}^{n_{max}}
    \qty(\mathcal{L}_{rec,n} + \mathcal{L}_{mask,n}).
    \label{eqn:mse-loss-with-mask-and-rec} 
\end{equation}
where $M_n$ represents the decoded mask associated with the $n$th token, $\odot$ is the Hadamard product and $D$ is the dimension of the matrix. In words, for each decoded mask and reconstruction, we only impose the MSE loss on regions of the image that are included in the mask $M_n$. Although imposing this loss slightly impairs the reconstruction performance, it helps us achieve a better balance between our twin goals of good image reconstruction and finding intepretable tokens (see \cref{fig:in-doc-examples}). The modified algorithm for our model is defined in \cref{alg:mask-alg}, trained on the loss \cref{eqn:mse-loss-with-mask-and-rec} ($\mathcal{S}$ refers to a model that produces a mask from the residual $X - \hat{X}_{n-1}$.). We term this variant of VLE `masked VLE', in contrast to `vanilla VLE'. We remark that both models are trained in an unsupervised fashion (more specifically, self-supervised in the case of the masked VLE model).

\begin{algorithm}
\caption{VLE Autoregressive Loop with Masks}\label{alg:mask-alg}
\begin{algorithmic}
\State $n \gets 0$
\State $\hat{X}_n \gets 0$    \Comment{Set the reconstruction to 0}
\State $\hat{M}_n \gets 0$    \Comment{Set the mask to 0}
\While{$n < n_{max}$}
    \State $\tilde{X}_n, \tilde{M}_n \gets \mathcal{S}(X - \hat{X}_{n-1})$ \Comment{Extract the mask to save for loss computation}
    \State $z_n \gets \mathcal{E}(\tilde{X}_n \mid\tilde{M}_n)$ \Comment{Encode the transformed image, conditioned on the mask}
    \State $X_n \gets \mathcal{D}(z_n)$
    \State $\hat{X}_n \gets \hat{X}_{n-1} + X_n$
    \State $\hat{M}_n \gets \hat{M}_{n-1} + \tilde{M}_n$
    \State $n \gets n + 1$
\EndWhile
\end{algorithmic}
\end{algorithm}

\section{Experiments}\label{sec:exper}

\subsection{Model Architecture}\label{sec:arch}
We experiment with a very simple autoencoder structure which we implement from decomposing a U-Net model by removing the skip-connections. Each layer consists of a $n$ residual blocks followed by a downsampling convolution layer. The dimension of the latent space is 64x smaller than the original image (same as the VAE we benchmark against). To iterate quickly on the experiments, we use a model with a small number of trainable parameters: $\sim$7M parameters for both the vanilla VLE and the mask VLE. 

\begin{figure}
    \centering
    \includegraphics[width=\textwidth]{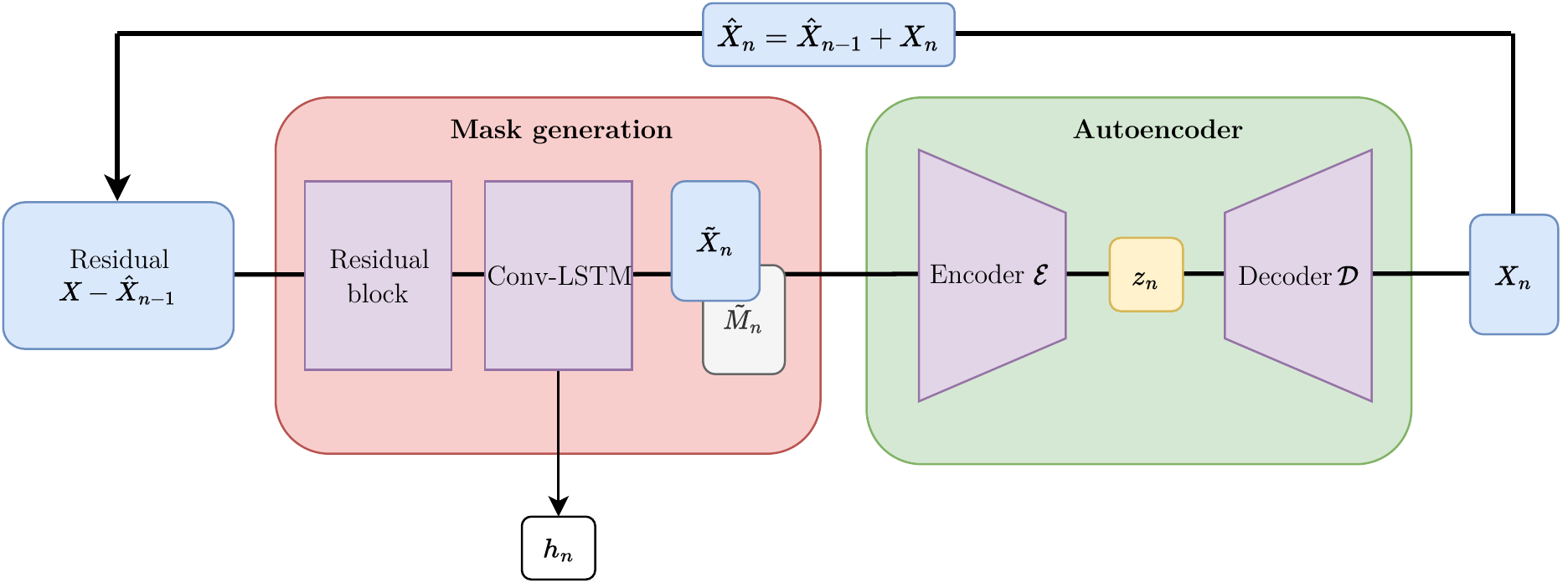}
    \caption{One iteration of \cref{alg:mask-alg}.}
    \label{fig:arch}
\end{figure}

For mask VLEs we include a ``precursor'' model we define in \cref{alg:mask-alg} as $\mathcal{S}$. The purpose of this model is to identify a focus object for the encoder to compress and is just a residual block. Additionally, we add a small convolutional LSTM after the precursor model of the encoder to provide a memory state for the encoder. This LSTM layer uses the hidden state to guide what objects to look at in relation to the previously seen objects in the image. For a schematic representation of our architecture, see \cref{fig:arch}. 

\subsection{Training}
We experimented with a number of different VLE variants, which comprised of small modifications to loss functions or model architecture (keeping intact the core idea of a variable length embedding). Notably, we find that different variants corresponded to different decompositions of the image. Whereas some models performed something as simple as a naive color decomposition, remarkably, other variants were able to identify meaningfully distinct objects in an image in an unsupervised manner.

During training, the minimum number of tokens is one in the vanilla case (which simply corresponds to a generic, fixed dimension autoencoder). In the mask case, we start with two tokens. The number of tokens used in a given iteration is chosen somewhat randomly. We sample the number of tokens from a folded-normal distribution where the mean value of the folded-normal distribution increases linearly with the number of iterations. This means at low iterations, the model should see smaller number of tokens and at higher iterations, the opposite. Five tokens was determined by the capacity of the GPUs' memory. We do observe that the model does generalize well when trained jointly and that an additional number of tokens used during inference can only boost reconstruction performance (as observed in \cref{fig:msextokens}).

We train on the LAION dataset \citep{laion}. The aspect ratios of the images vary, but we reshape them to have a fixed $512\times 512$ resolution. We parallelize each training run across 4 Nvidia A100 80GB SXM4 GPUs with a batch size of 16 (4 images/GPU). We train every model used in this paper to 120000 gradient updates (30000 steps/GPU).

\subsection{Evaluation}\label{sec:eval}

We evaluate the models on their reconstruction performance for 10000 images from a combination of iNaturalist \citep{inaturalist} and ImageNet \citep{imagenet} datasets and benchmark against the VAE used in Latent Diffusion Models \citep{rombach_high-resolution_2022}. We also note here again, as shown in \cref{table:eval-chart}, that the mask VLE does perform worse in reconstruction as a consequence of outputting more interpretable tokens compared to vanilla VLEs.

\begin{table}[h!]
\centering
\begin{tabular}{ |c||ccc| } 
\hline
 Token number & Vanilla VLE & Mask VLE & VAE \\
\hline
1 & 0.0119/0.645 & 0.2162/0.368 & 0.0113/0.654\\ 
2 & 0.0073/0.742 & 0.1355/0.272 & -\\ 
3 & 0.0053/0.804 & 0.1156/0.285 & -\\
4 & 0.0045/0.829 & 0.0098/0.656 & -\\ 
5 & 0.0041/0.841 & 0.0115/0.640 & -\\ 
6 & 0.0038/0.849 & 0.0099/0.655 & -\\ 
7 & 0.0036/0.854 & 0.0103/0.654 & -\\
\hline
\end{tabular}
\caption{Mean squared error and structural similarity index measure (SSIM) \citep{SSIM} for two variants of VLE (vanilla and mask), compared to VAE. The token dimension is downsampled by a factor of 64 from input dimension for both VLE and VAE. The \href{https://huggingface.co/stabilityai/sd-vae-ft-mse}{VAE architecture} has $\sim$80M parameters, while the VLE architecture is described in \cref{sec:arch} and has $\sim$7M parameters. We emphasize that we are able to achieve similar results to a pretrained VAE with less than a tenth of the parameters as seen in the first row between the vanilla VLE and the VAE.}
\label{table:eval-chart}
\end{table}

\section{Conclusion}
Motivated by an information-theoretic approach to compression and representation, we introduce a way to allow the latent dimension of an autoencoder to vary. We do this by representing images as (variable length) sequences of tokens, terming this technique Variable Length Embeddings. However, rather than purely aiming to compress the image as much as possible, we include an inductive bias that encourages the model to learn interpretable tokens via the masking mechanism discussed in \cref{sec:sem-pix}. We find that the model is then able to perform well on both tasks: it achieves a competitive reconstruction error compared to other autoencoders and finds decompositions of the image into human interpretable masks. 


Although we already found our models to output good masks, we believe their quality could be improved further by
\begin{itemize}
    \item adding image segmentation or saliency priors/semi-supervised losses so that the model can better understand the objects in the image, or
    \item adding other modalities (such as image captioning) during training so the model better understands the contextual objects in an image.
\end{itemize}
\paragraph{Extension to Generative Modeling} We believe VLEs have the potential to address a number of shortcomings for diffusion-based approaches, particularly the failure of diffusion models to accurately place objects in a user-specified quadrant.  Since each token in VLE fully characterizes particular objects in an image, they must contain information about their spatial location within the frame; this spatial information is much more natural to manipulate compared to diffusion-based approaches, for which it is not so clear where spatial information of individual objects is encoded. A first experiment in this direction would be to use a VLE model to encode all objects in a provided image and randomize the location of the objects without modifying the characteristics of these objects. Following this, one might  extend VLEs as an end-to-end text-to-image generative model. Text prompts to the model would be embedded as a sequence of tokens, with each token having a direct connection to some subject in the prompt. For instance, the prompt ``a penguin on the right side standing on a bed of ice with the Sun on the upper left side" would contain at least 3 key tokens: the penguin on the right, the bed of ice below, and the Sun on the upper-left side. \\

\paragraph{Acknowledgements} We thank key members of the Runway team that supported and provided guidance in our research: Rohan Agarwal, Jonathan Granskog, Deepti Ghadiyaram, Patrick Esser, Anastasis Germanidis. We additionally thank our friend Joe Zou for his valuable feedback on early drafts of this manuscript.



\bibliography{references}

\begin{thebibliography}{11}
\providecommand{\natexlab}[1]{#1}
\providecommand{\url}[1]{\texttt{#1}}
\expandafter\ifx\csname urlstyle\endcsname\relax
  \providecommand{\doi}[1]{doi: #1}\else
  \providecommand{\doi}{doi: \begingroup \urlstyle{rm}\Url}\fi

\bibitem[Horn et~al.(2018)Horn, Aodha, Song, Cui, Sun, Shepard, Adam, Perona,
  and Belongie]{inaturalist}
Grant~Van Horn, Oisin~Mac Aodha, Yang Song, Yin Cui, Chen Sun, Alex Shepard,
  Hartwig Adam, Pietro Perona, and Serge Belongie.
\newblock The inaturalist species classification and detection dataset, 2018.

\bibitem[Deng et~al.(2009)Deng, Dong, Socher, Li, Li, and Fei-Fei]{imagenet}
Jia Deng, Wei Dong, Richard Socher, Li-Jia Li, Kai Li, and Li~Fei-Fei.
\newblock Imagenet: A large-scale hierarchical image database.
\newblock In \emph{2009 IEEE Conference on Computer Vision and Pattern
  Recognition}, pages 248--255, 2009.
\newblock \doi{10.1109/CVPR.2009.5206848}.

\bibitem[Mallat and Zhang(1993)]{pursuit1993}
S.G. Mallat and Zhifeng Zhang.
\newblock Matching pursuits with time-frequency dictionaries.
\newblock \emph{IEEE Transactions on Signal Processing}, 41\penalty0
  (12):\penalty0 3397--3415, 1993.
\newblock \doi{10.1109/78.258082}.

\bibitem[Toderici et~al.(2017)Toderici, Vincent, Johnston, Hwang, Minnen, Shor,
  and Covell]{toderici_full_2017}
George Toderici, Damien Vincent, Nick Johnston, Sung~Jin Hwang, David Minnen,
  Joel Shor, and Michele Covell.
\newblock Full {Resolution} {Image} {Compression} with {Recurrent} {Neural}
  {Networks}.
\newblock In \emph{2017 {IEEE} {Conference} on {Computer} {Vision} and
  {Pattern} {Recognition} ({CVPR})}, pages 5435--5443, Honolulu, HI, July 2017.
  IEEE.
\newblock ISBN 9781538604571.
\newblock \doi{10.1109/CVPR.2017.577}.
\newblock URL \url{http://ieeexplore.ieee.org/document/8100060/}.

\bibitem[Gregor et~al.(2015)Gregor, Danihelka, Graves, Rezende, and
  Wierstra]{pmlr-v37-gregor15}
Karol Gregor, Ivo Danihelka, Alex Graves, Danilo Rezende, and Daan Wierstra.
\newblock Draw: A recurrent neural network for image generation.
\newblock In Francis Bach and David Blei, editors, \emph{Proceedings of the
  32nd International Conference on Machine Learning}, volume~37 of
  \emph{Proceedings of Machine Learning Research}, pages 1462--1471, Lille,
  France, 07--09 Jul 2015. PMLR.
\newblock URL \url{https://proceedings.mlr.press/v37/gregor15.html}.

\bibitem[Ho et~al.(2020)Ho, Jain, and Abbeel]{ho2020denoising}
Jonathan Ho, Ajay Jain, and Pieter Abbeel.
\newblock Denoising diffusion probabilistic models.
\newblock \emph{Advances in Neural Information Processing Systems},
  33:\penalty0 6840--6851, 2020.

\bibitem[Zhang et~al.(2018)Zhang, Isola, Efros, Shechtman, and Wang]{zhang2018}
Richard Zhang, Phillip Isola, Alexei~A. Efros, Eli Shechtman, and Oliver Wang.
\newblock The unreasonable effectiveness of deep features as a perceptual
  metric, 2018.

\bibitem[van~der Walt et~al.(2014)van~der Walt, {S}ch\"onberger,
  {Nunez-Iglesias}, {B}oulogne, {W}arner, {Y}ager, {G}ouillart, {Y}u, and the
  scikit-image contributors]{scikit-image}
{S}t\'efan van~der Walt, {J}ohannes~{L}. {S}ch\"onberger, {J}uan
  {Nunez-Iglesias}, {F}ran\c{c}ois {B}oulogne, {J}oshua~{D}. {W}arner, {N}eil
  {Y}ager, {E}mmanuelle {G}ouillart, {T}ony {Y}u, and the scikit-image
  contributors.
\newblock scikit-image: image processing in {P}ython.
\newblock \emph{PeerJ}, 2:\penalty0 e453, 6 2014.
\newblock ISSN 2167-8359.
\newblock \doi{10.7717/peerj.453}.
\newblock URL \url{https://doi.org/10.7717/peerj.453}.

\bibitem[Schuhmann et~al.(2021)Schuhmann, Vencu, Beaumont, Kaczmarczyk, Mullis,
  Katta, Coombes, Jitsev, and Komatsuzaki]{laion}
Christoph Schuhmann, Richard Vencu, Romain Beaumont, Robert Kaczmarczyk,
  Clayton Mullis, Aarush Katta, Theo Coombes, Jenia Jitsev, and Aran
  Komatsuzaki.
\newblock Laion-400m: Open dataset of clip-filtered 400 million image-text
  pairs, 2021.

\bibitem[Rombach et~al.(2022)Rombach, Blattmann, Lorenz, Esser, and
  Ommer]{rombach_high-resolution_2022}
Robin Rombach, Andreas Blattmann, Dominik Lorenz, Patrick Esser, and Bjorn
  Ommer.
\newblock High-{Resolution} {Image} {Synthesis} with {Latent} {Diffusion}
  {Models}.
\newblock In \emph{2022 {IEEE}/{CVF} {Conference} on {Computer} {Vision} and
  {Pattern} {Recognition} ({CVPR})}, pages 10674--10685, New Orleans, LA, USA,
  June 2022. IEEE.
\newblock ISBN 9781665469463.
\newblock \doi{10.1109/CVPR52688.2022.01042}.
\newblock URL \url{https://ieeexplore.ieee.org/document/9878449/}.

\bibitem[Wang et~al.(2004)Wang, Bovik, Sheikh, and Simoncelli]{SSIM}
Zhou Wang, A.C. Bovik, H.R. Sheikh, and E.P. Simoncelli.
\newblock Image quality assessment: from error visibility to structural
  similarity.
\newblock \emph{IEEE Transactions on Image Processing}, 13\penalty0
  (4):\penalty0 600--612, 2004.
\newblock \doi{10.1109/TIP.2003.819861}.

\end{thebibliography}

\appendix
\section{Masked VLE Results}\label{sec:mask-vae-res}
Below, we show a fair sample of the results for a masked VLE model with $\sim$7M parameters trained for $\sim$120,000 gradient updates. For further details see \cref{sec:exper}.
\begin{figure}[H]
    \centering
    \includegraphics[width=0.93\textwidth]{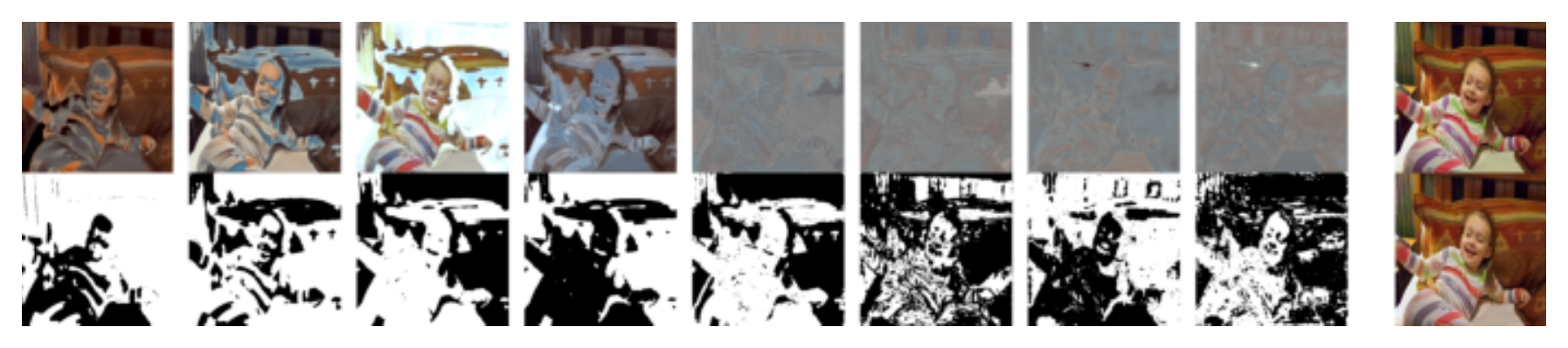}
    \includegraphics[width=0.93\textwidth]{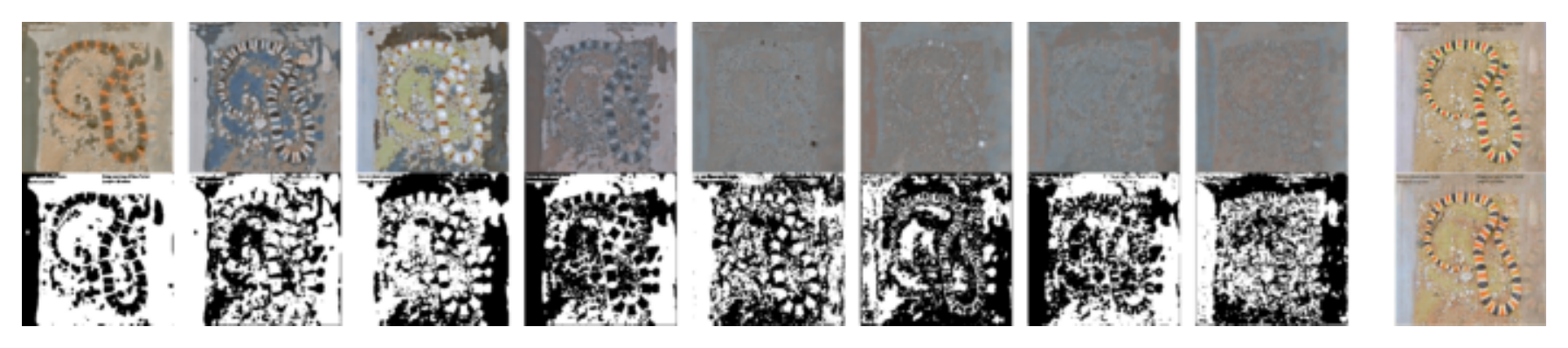}
    \includegraphics[width=0.93\textwidth]{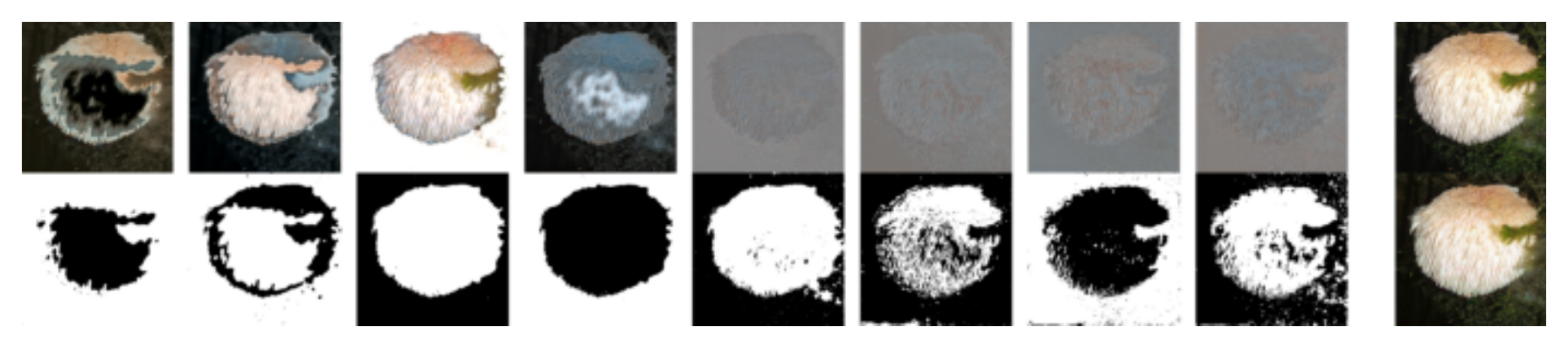}
\end{figure}

\begin{figure}[H]
    \centering
    \includegraphics[width=0.93\textwidth]{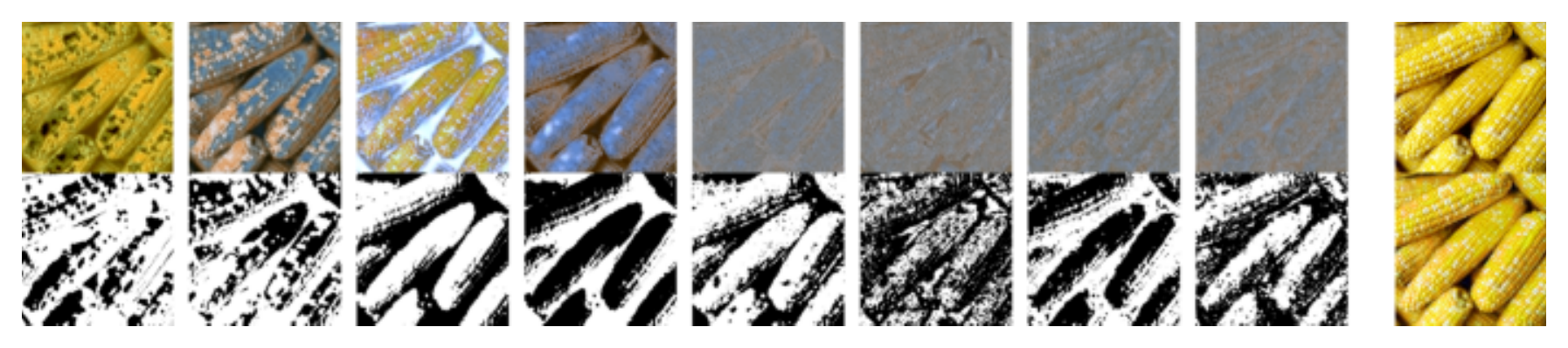}
    \includegraphics[width=0.93\textwidth]{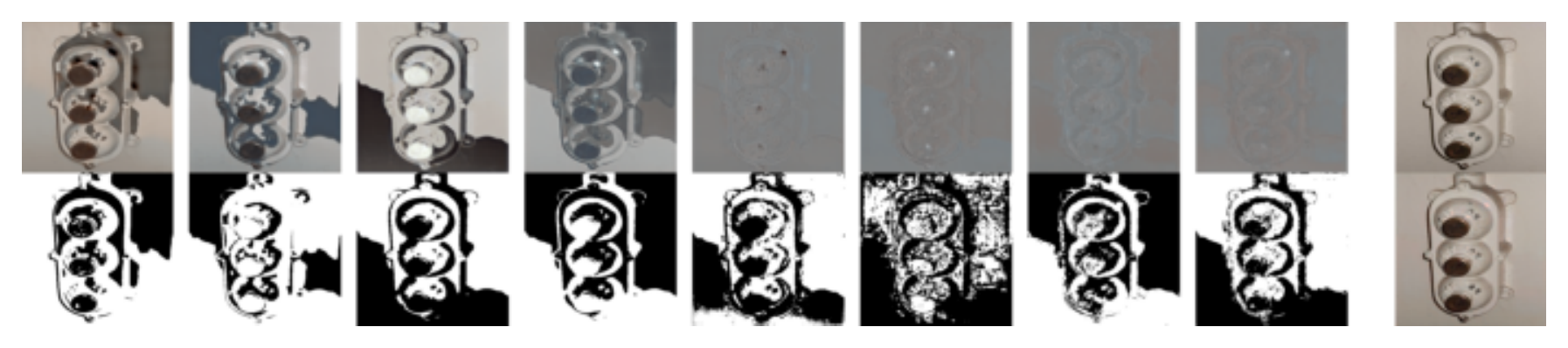}
    \includegraphics[width=0.93\textwidth]{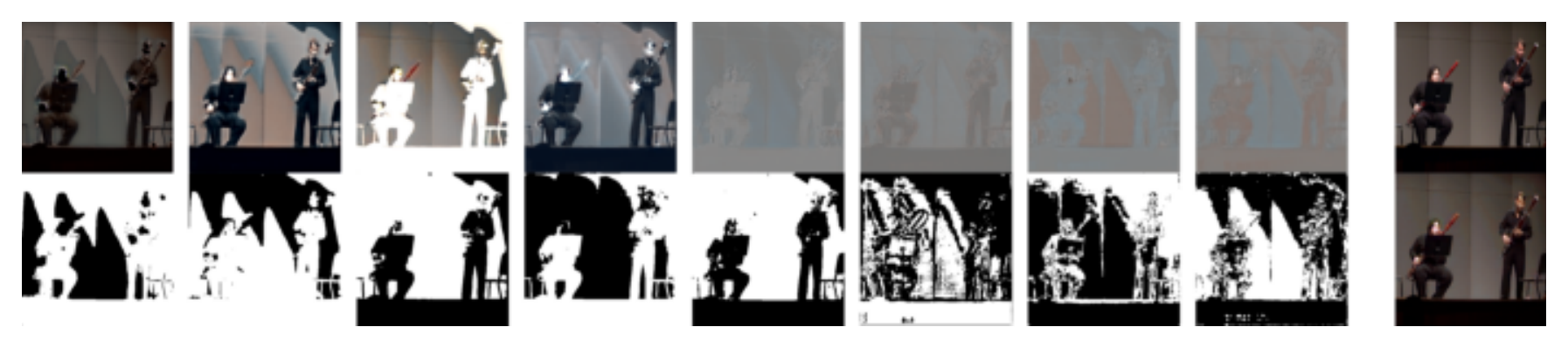}
    \includegraphics[width=0.93\textwidth]{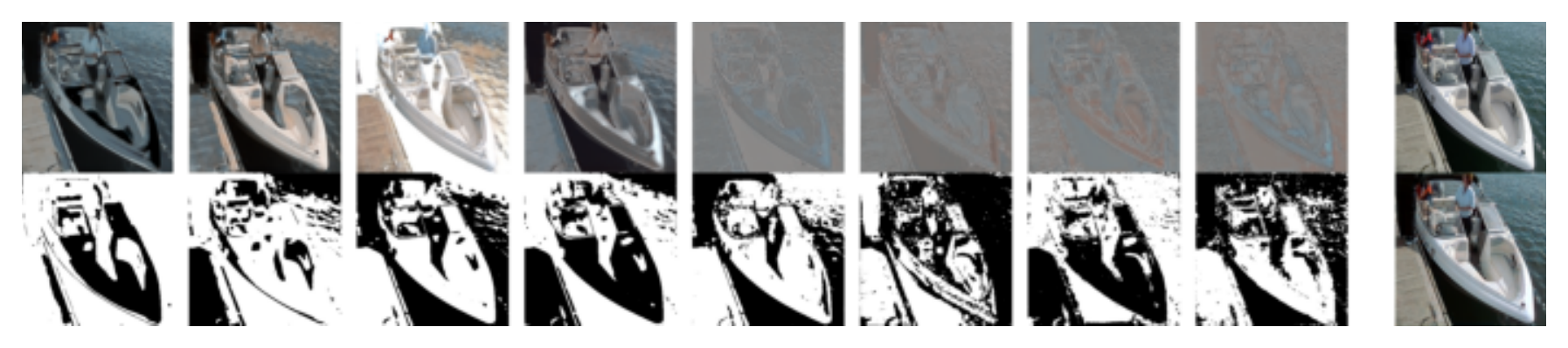}    
    \includegraphics[width=0.93\textwidth]{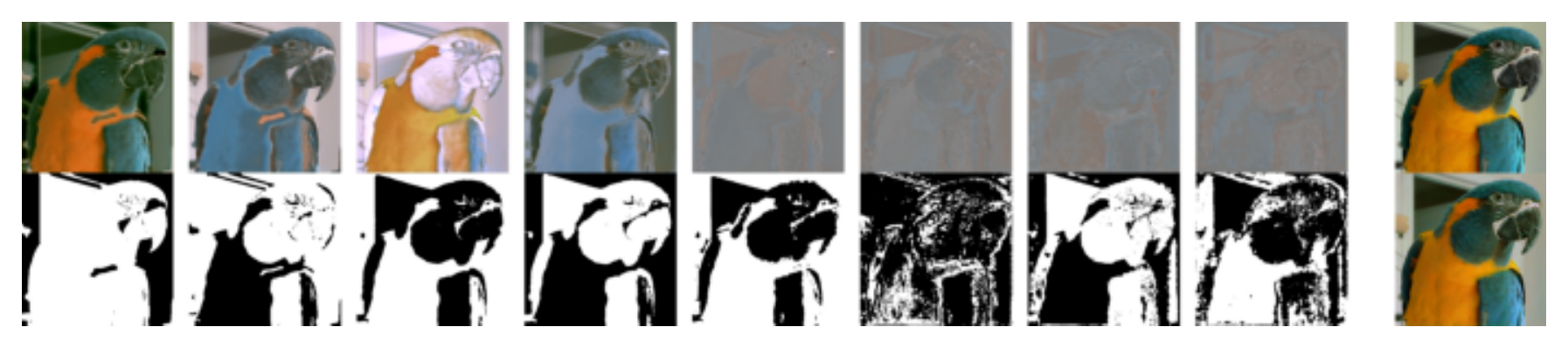}
    \includegraphics[width=0.93\textwidth]{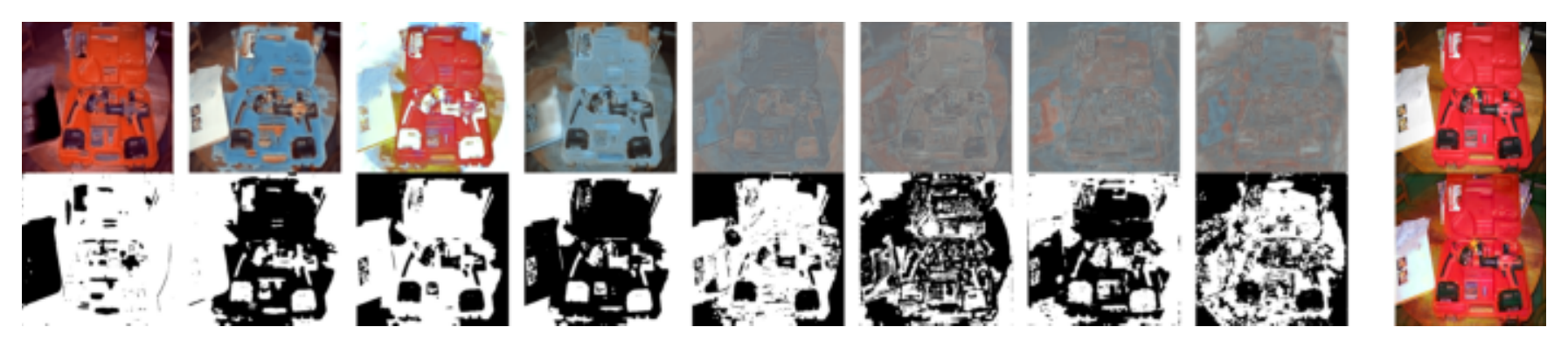}
    \caption{Randomly selected samples of the intermediate reconstructions (top row) and corresponding masks (bottom row) using 8 tokens. The final column on the right represents the source (top image) and reconstruction (bottom image).}
    \label{fig:mask_examples}
\end{figure}

\clearpage
\section{Vanilla VLE Results}\label{sec:vanilla-vae-res}
Below, we show a fair sample of the results for a vanilla VLE model.
\begin{figure}[H]
    \centering
    \includegraphics[width=0.93\textwidth]{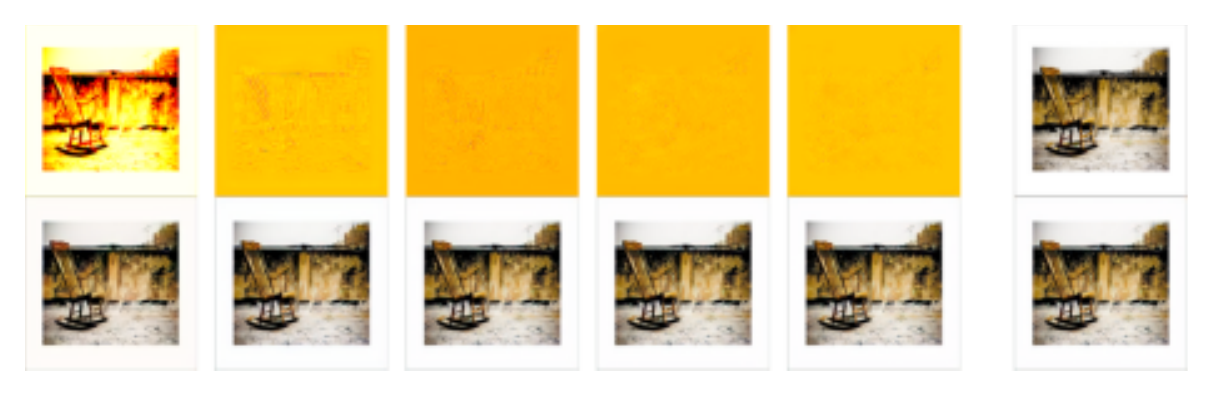}
    \includegraphics[width=0.93\textwidth]{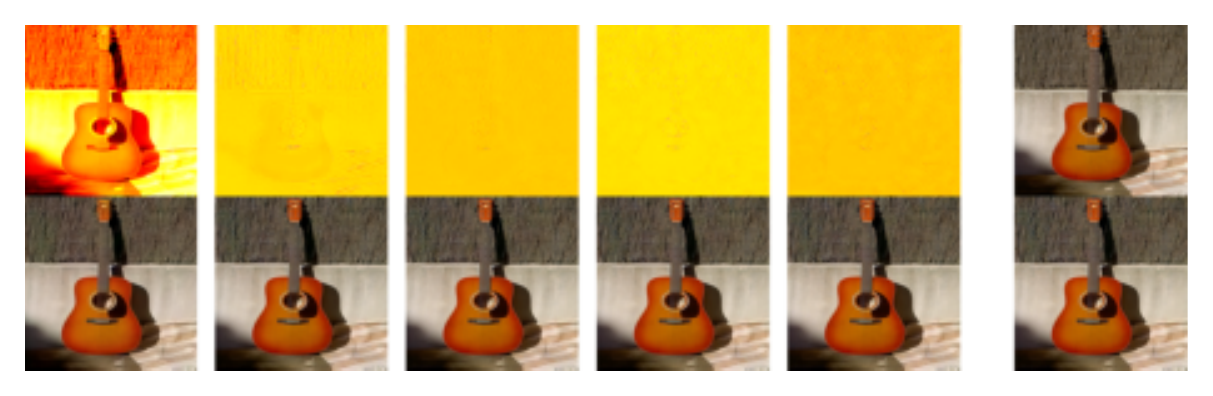}
    \includegraphics[width=0.93\textwidth]{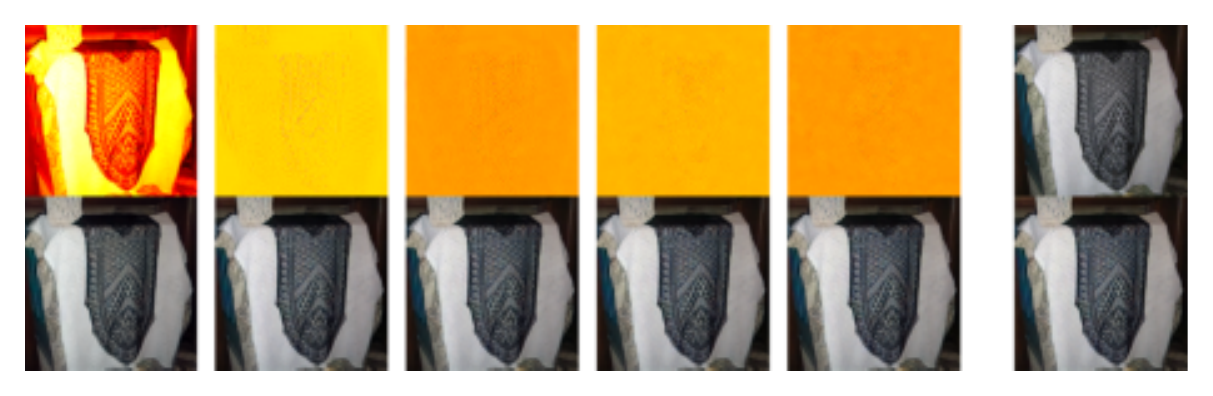}   
    \includegraphics[width=.93\textwidth]{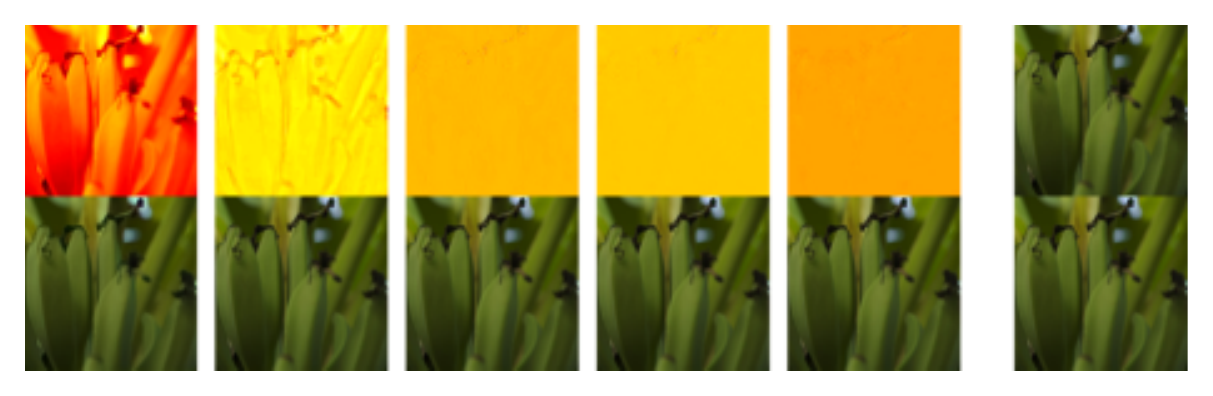}
\end{figure}

\begin{figure}[H]
    \centering
    \includegraphics[width=0.93\textwidth]{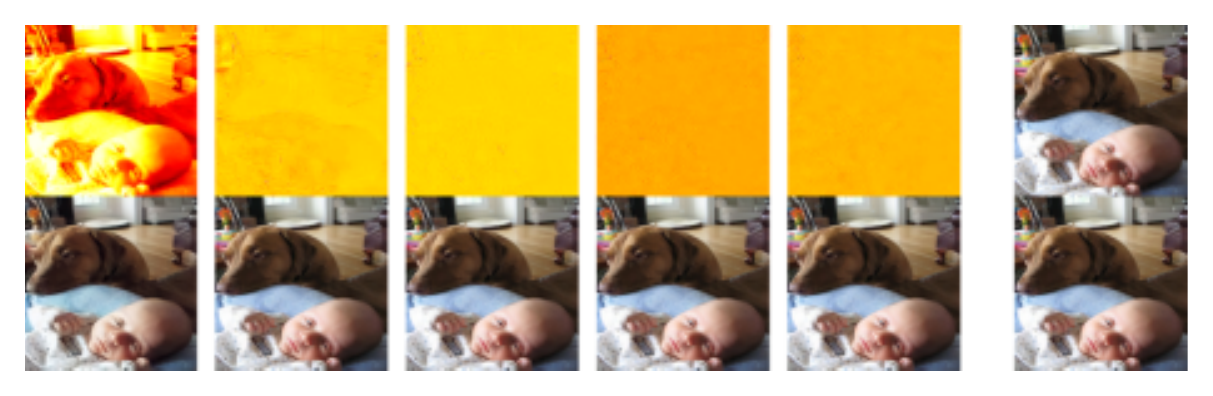}
    \includegraphics[width=0.93\textwidth]{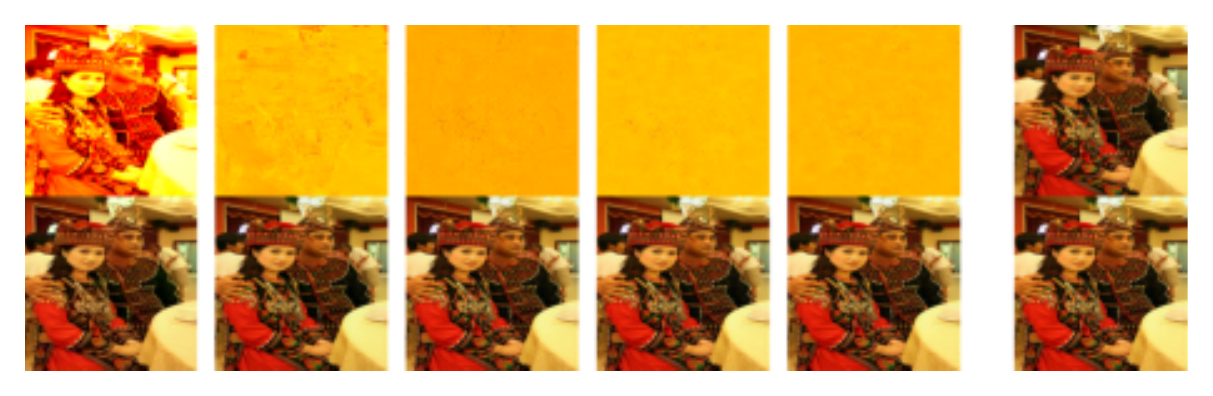}
    \includegraphics[width=0.93\textwidth]{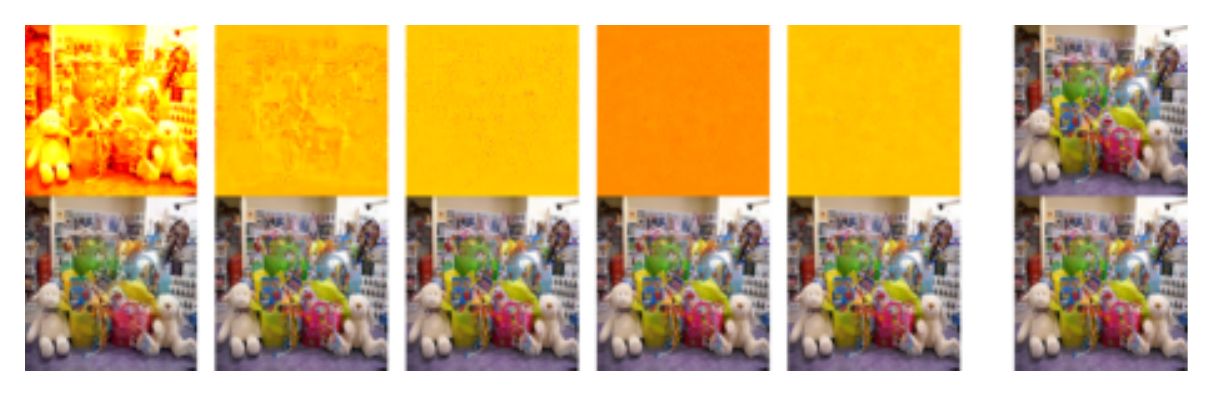}
    \caption{Randomly selected samples of the intermediate reconstructions in heatmap space (top row) and sum of the previous reconstructions in RGB space (bottom row) using 5 tokens. The latent space we used for the vanilla VLE captures the majority of the image but looking closely, one can see how the model adds higher frequency detail in later tokens. The final column on the right represents the source (top image) and reconstruction (bottom image).}
    \label{fig:vanilla_examples}
\end{figure}

\end{document}